\documentclass{article}

\usepackage[english]{babel}
\usepackage[letterpaper,top=2cm,bottom=2cm,left=2cm,right=2cm,marginparwidth=1.5cm]{geometry}
\usepackage[utf8]{inputenc}
\usepackage[T1]{fontenc}
\usepackage{hyperref}
\usepackage{url}
\usepackage{booktabs}
\usepackage{amsfonts}
\usepackage{nicefrac}
\usepackage{microtype}
\usepackage{xcolor}
\usepackage{multirow}
\usepackage{amsmath}
\usepackage{enumitem}
\usepackage{graphicx}
\usepackage{colortbl}
\usepackage{wrapfig}
\usepackage{pifont}
\usepackage{natbib}
\usepackage{amsthm}
\usepackage[table]{xcolor}
\usepackage{arydshln}
\usepackage{tabularx}
\usepackage{makecell}
\usepackage{algorithm}
\usepackage{algpseudocode}
\theoremstyle{plain}
\newtheorem{theorem}{Theorem}[section]
\newtheorem{proposition}[theorem]{Proposition}
\newtheorem{lemma}[theorem]{Lemma}
\usepackage{threeparttable}

\usepackage[most]{tcolorbox}
\usepackage{xcolor}
\newtcolorbox{takeawaybox}{
  colback=gray!10,
  colframe=gray!10,
  boxrule=0pt,
  arc=3mm,
  left=3mm,
  right=3mm,
  top=2mm,
  bottom=2mm,
  width=\linewidth,
}

\title{\emph{SCOPE}: Structured Prototype-Guided Adaptation for EEG Foundation Models with Limited Labels}

\begin{document}

\author{
    Jingying Ma\textsuperscript{1}\thanks{Co-first Authors.},
    Feng Wu\textsuperscript{1}\textsuperscript{*},
    Yucheng Xing\textsuperscript{1,3},
    Qika Lin\textsuperscript{1},
    Tianyu Liu\textsuperscript{1,3}, \\
    Chenyu Liu\textsuperscript{4}\thanks{Corresponding Authors: \texttt{chenyu003@e.ntu.edu.sg, jia.ziyu@outlook.com}.},
    Ziyu Jia\textsuperscript{5,6}\textsuperscript{†},
    Mengling Feng\textsuperscript{1,2}
}
\date{}

\newcommand{\affiliations}{
    \textsuperscript{1}Saw Swee Hock School of Public Health, National University of Singapore, Singapore \\
    \textsuperscript{2}Institute of Data Science, National University of Singapore, Singapore \\
    \textsuperscript{3}Guangzhou Research Translation and Innovation Institute, \\National University of Singapore, Guangzhou, China\\
    \textsuperscript{4}College of Computing and Data Science, Nanyang Technological University, Singapore\\
    \textsuperscript{5}Beijing Key Laboratory of Brainnetome and Brain-Computer Interface, \\Institute of Automation, Chinese Academy of Sciences, Beijing, China\\
    \textsuperscript{6}Brainnetome Center, Institute of Automation, Chinese Academy of Sciences, Beijing, China\\
    }

\maketitle
{\centering\affiliations\par}

\begin{abstract}
  Electroencephalography (EEG) foundation models (EFMs) have achieved strong performance under full fine-tuning but exhibit poor generalization when subject-level supervision is limited, a common constraint in real-world clinical settings. We show that this failure stems not merely from limited supervision, but from \textit{a structural mismatch between noisy, limited supervision and the highly plastic parameter space of EFMs}. To address this challenge, we propose \emph{SCOPE}, a Structured COnfidence-aware Prototype-guided adaptation framework for EFM fine-tuning. \emph{SCOPE} follows a two-stage pipeline. In the first stage, we construct reliable external supervision by learning geometry-regularized task priors, constructing balanced class-level prototypes over the resulting embeddings, and producing confidence-aware pseudo-labels from their agreement to filter unreliable signals on unlabeled data. In the second stage, we introduce \emph{ProAdapter}, which adapts frozen EEG foundation models via a lightweight adapter conditioned on the structured prototypes. Experiments across three EEG tasks and five foundation model backbones demonstrate that \emph{SCOPE} consistently achieves strong performance and efficiency under label-limited cross-subject settings.
\end{abstract}

\section{Introduction}
\label{intro}

Electroencephalography (EEG) is widely used to characterize brain states across clinical contexts \citep{mushtaq2024one}, enabling sleep staging\citep{shuai2026osf}, affective assessment\citep{sun2024adaptive}, and noninvasive brain-computer interfaces \citep{ding2025eeg}. Recent research has increasingly shifted from task-specific models \citep{liu2024vbh, ma2025st} toward EEG foundation models (EFMs), which leverage large-scale pretraining to learn transferable representations across tasks and datasets \citep{zhou2025brain, wang2026deeperbrain, huang2026unified}. In experimental settings, EFMs achieve strong performance via full fine-tuning, making it the dominant adaptation strategy in current EFM research \citep{jianglarge, wangcbramod, zhou2025csbrain, ma2025codebrain}.

\begin{figure}[t!]
    \centering
    \includegraphics[width=1\linewidth]{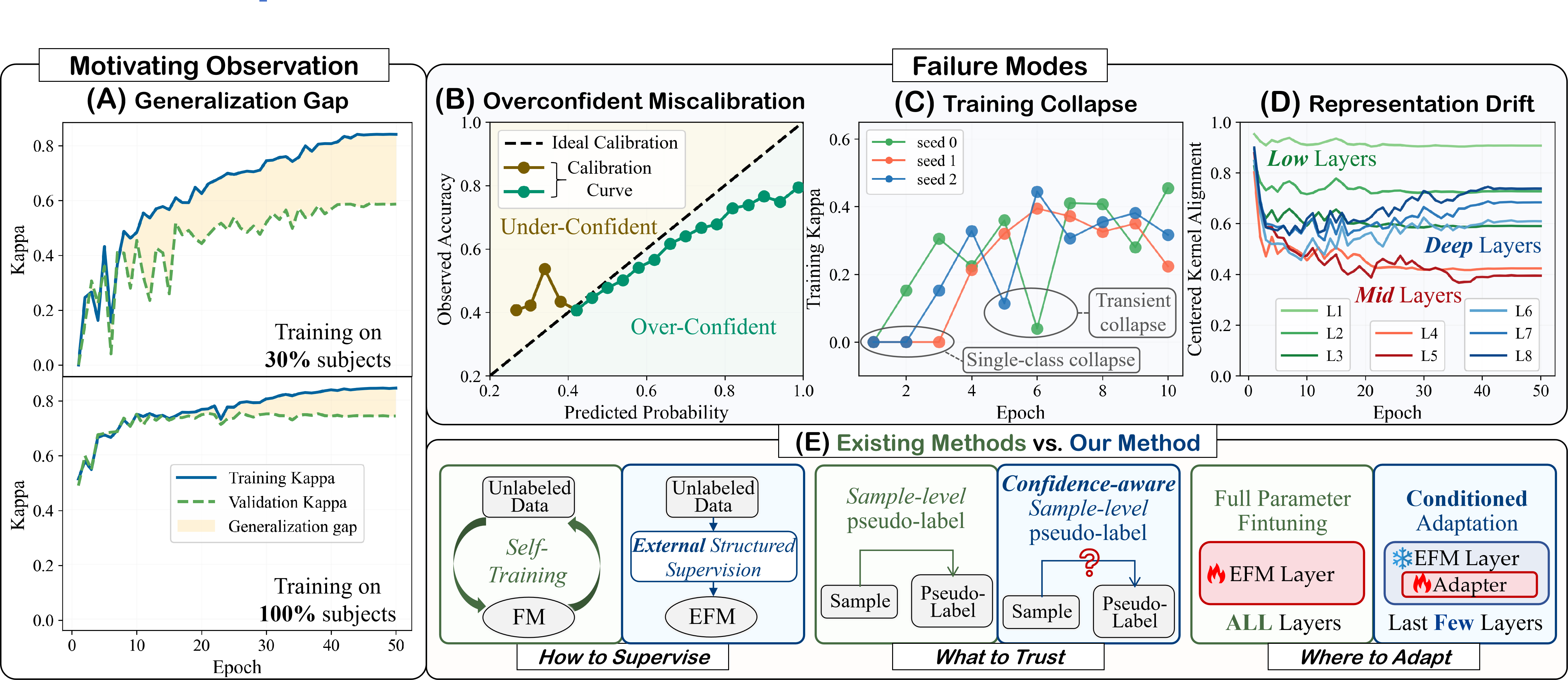}
    \caption{Empirical motivation for label-limited EFM adaptation: adapting \emph{CodeBrain} on ISRUC with 30\% labeled subjects reveals 
    (A) large generalization gap;
    (B) overconfident miscalibration;
    (C) training collapse across seeds; and
    (D) layer-wise representation drift.
    (E) These observations motivate our shift from self-training, sample-level pseudo-labels, and full fine-tuning toward external cohort-level structured supervision, confidence-aware pseudo-labeling, and conditioned adaptation.
    \label{fig:intro}}
\end{figure}

However, real-world EEG adaptation typically involves only a few labeled subjects alongside abundant unlabeled recordings, since reliable labels require expert subject-level annotation \citep{fiorillo2023u}. Despite strong benchmark performance, \textbf{EFMs exhibit clear generalization degradation under full fine-tuning in this label-limited regime}: as shown in Figure~\ref{fig:intro}(A), training on 30\% of subjects produces a substantially larger generalization gap than full supervision \citep{lea2024mind, ma2025development}. A closer look reveals that 1) the training kappa still converges to a level comparable to full supervision, and 2) the gap continues to widen with further training. This indicates that the capacity is more than sufficient and that additional optimization merely overfits the limited supervision rather than consolidating transferable representations. Therefore, the unreliability of full fine-tuning reflects \textbf{\textit{a structural mismatch between limited supervision and the large, highly plastic parameter space of EFMs}}, motivating us to revisit the problem of adapting EFMs under label-limited conditions, where several challenges are identified.

\textbf{1) EFMs fail to provide reliable self-guidance.}
Self-training is widely used to leverage unlabeled data under limited supervision in foundation models \citep{xie2020self, zhang2025revisiting}, relying on the assumption that model confidence correlates with prediction accuracy \citep{sohn2020fixmatch, berthelotremixmatch}. However, as evidenced by Figure~\ref{fig:intro}(B), EFMs are overconfident but inaccurate under limited supervision. In this regime, miscalibrated EFM predictions make self-generated supervision potentially misleading and prone to error amplification.

\textbf{2) EFMs are prone to prediction collapse.}
Pseudo-labeling is commonly used to construct supervision from unlabeled data \citep{zhang2021flexmatch, wangfreematch}. However, properties of EEG data make pseudo-labeling easily biased: continuous brain states induce ambiguous class boundaries \citep{zhou2025continuous}, while class imbalance and intra-class multimodality bias learning toward dominant modes \citep{berry2017aasm, rizvedefense, guo2022class}. Unfortunately, EFMs are particularly sensitive, as evidenced by the prediction collapse under limited supervision (Figure~\ref{fig:intro}(C)).

\textbf{3) EFMs' parameter space is vulnerable.}
EFMs rely on a large pretrained parameter space to encode transferable representations \citep{yang2025foundation}. Under limited supervision, full fine-tuning leads to large degradation in representation similarity across layers (Figure~\ref{fig:intro}(D)). This reveals that noisy supervision-driven gradients can easily diffuse across layers, overwriting transferable pretrained representations and inducing irreversible optimization drift \citep{kumarfine, ding2023parameter, zhang2024dissecting}.

To address the above challenges, we propose \emph{SCOPE}, a \underline{\textbf{S}}tructured \underline{\textbf{CO}}nfidence-aware \underline{\textbf{P}}rototype-guided adaptation framework for \underline{\textbf{E}}EG foundation model. To the best of our knowledge, \textbf{this is the first framework that systematically studies how to adapt EEG foundation models under label-limited practical deployment.} \textit{First}, we introduce an \emph{external supervision construction} module that builds cohort-level structured prototypes to provide persistent guidance. \textit{Second}, within the same module, we derive confidence-aware pseudo-labels to provide reliable sample-level supervision. \textit{Third}, we propose \emph{ProAdapter} to inject learned prototypes into deeper EFM layers via lightweight modulation while freezing the backbone to preserve pretrained knowledge. Our contributions are:

\begin{itemize}[leftmargin=*, itemsep=0em, topsep=0em]
    \item \textbf{External Cohort-level Structured Prototypes.} We construct cohort-level prototypes with prior-guided inter-class geometry \citep{papyan2020prevalence} and balanced intra-class structure \citep{sinkhorn1967concerning, caron2020unsupervised}, providing persistent external guidance for EFMs' adaptation.
    \item \textbf{Confidence-aware Sample-level Pseudo-labeling.} We derive externally guided pseudo-labels by measuring the agreement between two complementary geometric predictions, filtering out uncertain samples for reliable sample-level supervision.
    \item \textbf{Prototype-conditioned Adaptation.} We propose \emph{ProAdapter} to align EFM adaptation with structured external supervision by injecting learned prototypes through lightweight modulation.
    \item \textbf{Comprehensive Evaluation}. We conduct a broad evaluation across \textbf{50 label-limited adaptation settings}, spanning 6 EEG tasks, 5 EFM backbones, and labeled-subject ratios from 5\% to 50\%, showing that \emph{SCOPE} consistently delivers strong performance with limited trainable parameters.
\end{itemize}

\section{Related Work}
\label{related work}

\paragraph{Scalp EEG Foundation Models.} 
Self-supervised pretraining has been widely used to learn transferable EEG representations, including contrastive \citep{kostas2021bendr, karantonis2025subject} and reconstruction-based methods \citep{chien2022maeeg, mohammadi2024eeg2rep}. Much of the work focuses on scalp EEG, enabled by standardized benchmarks and data availability \citep{wangcbramod, obeid2016temple, xiong2025eeg}. Existing EFMs mainly adopt sequence modeling backbones, including Transformer family \citep{yang2023biot, jianglarge, wang2024eegpt, zhou2025csbrain} and state-space models (SSSMs) \citep{wang2025eegmamba, ma2025codebrain}. Although EFMs are powerful, current practice relies on full fine-tuning, which is suboptimal under label-limited adaptation \citep{lee2025large, yang2025foundation, wang2026benchmarking}.

\paragraph{Adaptation of EEG foundation models.}
Adapting EFMs under limited supervision requires constraining updates to preserve the pretrained knowledge. Useful design principles can be drawn from parameter-efficient fine-tuning (PEFT) in language and vision foundation models, spanning weight-space approaches that confine updates to low-dimensional subspaces \citep{hu2022lora, liu2022few, zhang2023adaptive}, feature-space modulation \citep{perez2018film, MahabadiR0H20}, and adapter-based techniques \citep{houlsby2019parameter, karimi2021compacter}. Related efforts in EFMs have pursued either parameter efficiency \citep{jeon2025parameter, choi2026pearl} or data efficiency via informative sample selection \citep{liao2026eegtune}. However, how to adapt EFMs under cross-subject, label-limited conditions remains underexplored.

\section{Methodology}
\label{method}

\subsection{Problem Formulation and Framework Overview}
\textbf{Problem Formulation.} Let $\mathbf{X} \in \mathbb{R}^{C \times T}$ denote an EEG signal with $C$ channels and length $T$. The training data comprise a small labeled set $\mathcal{D}_l = \{(\mathbf{X}_i, y_i)\}_{i=1}^{N_l}$, where $y_i \in \{1, \dots, K\}$ is the label, and a much larger unlabeled set $\mathcal{D}_u = \{\mathbf{X}_j\}_{j=1}^{N_u}$. Both sets share the same label space but exhibit distribution shifts due to inter-subject variability.

\begin{figure*}[t!]
    \centering
    \includegraphics[width=1\linewidth]{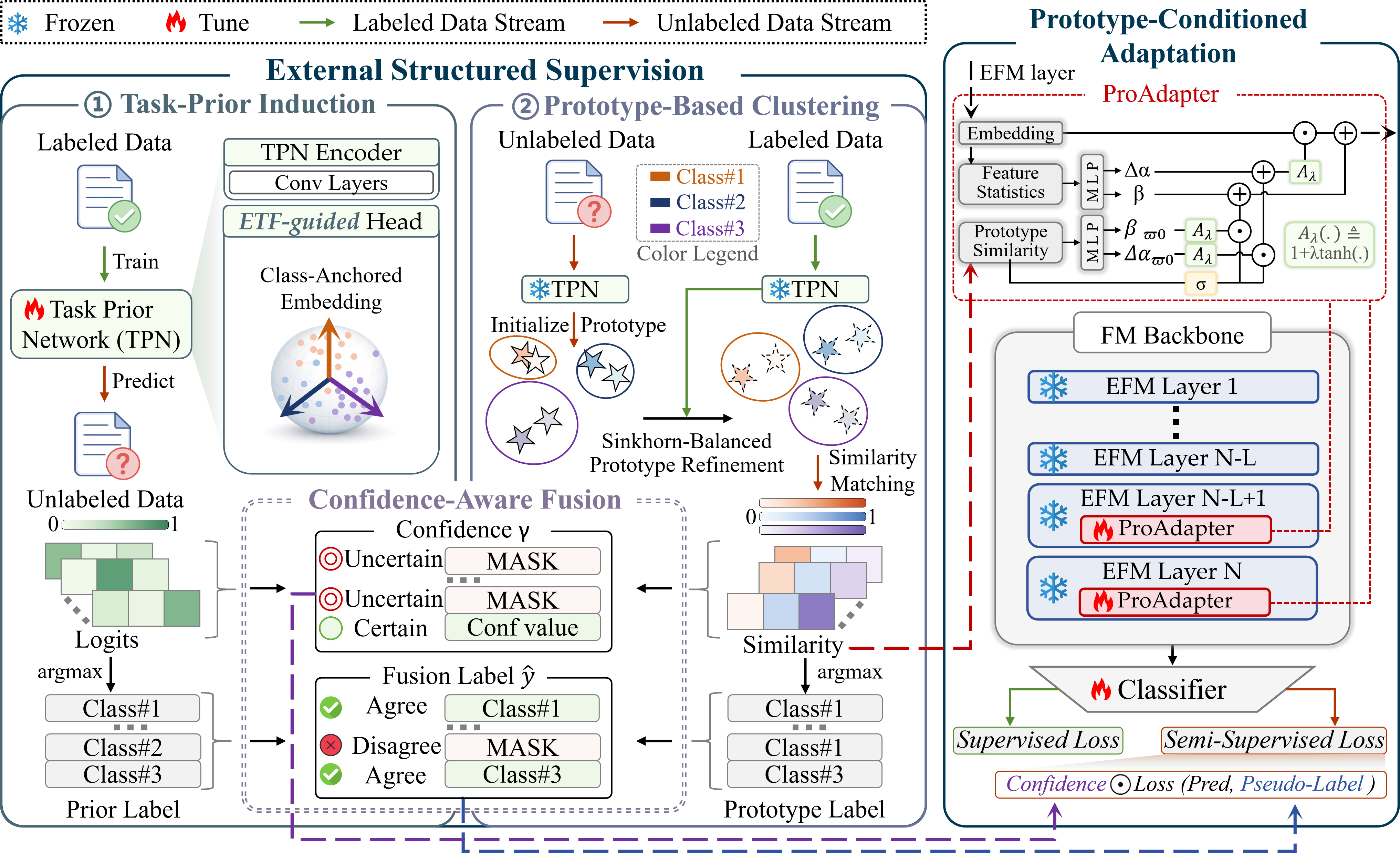}
    \caption{Overview of the two-stage \emph{SCOPE} framework. \textbf{Left}: External structured supervision construction progressively induces class-level prototypes and confidence-aware pseudo-labels for unlabeled data. \textbf{Right}: A frozen EEG foundation model is adapted via lightweight prototype-conditioned adapters in the last layers, using confidence-weighted supervision for controlled adaptation.}
    \label{fig:model}
\end{figure*}

\textbf{Framework Overview.} 
Motivated by the challenges in Figure~\ref{fig:intro}, \emph{SCOPE} follows a two-stage design: \emph{(I) external supervision construction} and \emph{(II) prototype-conditioned adaptation} (Figure~\ref{fig:model}). \emph{Stage I} addresses \textbf{C1 (EFMs fail to provide reliable self-guidance)} by constructing prototypes as persistent external structured supervision in Section~\ref{sec:structured_prototype}; it further addresses \textbf{C2 (EFMs are prone to prediction collapse)} by filtering out 
untrustworthy pseudo-labels through confidence-aware selection in Section~\ref{sec:confidence_fusion}. \emph{Stage II} addresses \textbf{C3 (EFMs' parameter space is vulnerable)} by fully freezing the EFM and modulating only its deep layers conditioned on prototypes in Section~\ref{sec:adapter}. Section~\ref{sec:objective} then presents 
the training strategy and overall objective.

\subsection{External structured supervision construction}
\label{sec:pseudo}
To provide EFMs with reliable supervision, we progressively construct cohort-level structured prototypes and confidence-aware sample-level pseudo-labels externally.

\subsubsection{Cohort-level Structured Prototype Construction}
\label{sec:structured_prototype}

\textit{\textbf{Technical Challenge.}}
Prototypes need to be discriminative across classes while preserving fine-grained intra-class structure. However, ambiguous EEG boundaries and imbalanced intra-class modes make simple prototype construction prone to class mixing and dominant-mode collapse.
\\
\textit{\textbf{Design Rationale.}}
We address this with hierarchical geometry. We first impose inter-class geometric regularization to obtain class-separable, cluster-friendly task-prior embeddings, and then learn multiple balanced prototypes per class to represent diverse intra-class patterns.

\textbf{Task-Prior Induction via Inter-Class Geometry.}
A lightweight task-prior network (TPN) $h_{\psi}$ is trained on the labeled subset $\mathcal{D}_l$,
where $h_{\psi}$ consists of a feature encoder $f_{\psi}$ and a classification head. Given an input EEG sample $\mathbf{X}_i$, the encoder produces a task-specific embedding
\begin{equation}
\mathbf{z}_i = f_{\psi}(\mathbf{X}_i)
= \mathcal{H}_{\mathrm{sep}}
\big(
\mathcal{H}_{\mathrm{spa}}
(
\mathcal{H}_{\mathrm{temp}}(\mathbf{X}_i)
)
\big),
\end{equation}
where $\mathcal{H}_{\mathrm{temp}}(\cdot)$, $\mathcal{H}_{\mathrm{spa}}(\cdot)$, and $\mathcal{H}_{\mathrm{sep}}(\cdot)$ are a temporal, depthwise spatial, and separable convolution block, respectively. For multi-class tasks, we impose an Equiangular Tight Frame (ETF)-based geometric regularization to enforce maximal inter-class angular separation.

\begin{proposition} 
\label{prop:etf_angle} 
Let $\{\tilde{\mathbf{w}}_{k}\}_{k=1}^{K}\subset\mathbb{S}^{d-1}$ be the normalized classifier weights. If they satisfy the simplex equiangular tight frame condition $ \tilde{\mathbf{w}}_{k}^{\top}\tilde{\mathbf{w}}_{k'} = -\frac{1}{K-1}, \qquad \forall\, k\neq k', $, the minimum pairwise angular separation is maximized.
\end{proposition}

Proposition~\ref{prop:etf_angle} provides theoretical motivation for explicit ETF regularization, which yields  uniformly separated class-wise embeddings under cross-entropy optimization. Specifically, let $W=[\mathbf{w}_{1},\dots,\mathbf{w}_{K}]\in\mathbb{R}^{d\times K}$ denote the weight matrix of the linear head.
We normalize each class weight as $\tilde{\mathbf{w}}_{k}=\mathbf{w}_{k}/\|\mathbf{w}_{k}\|_2$ to obtain $\tilde W = [\tilde{\mathbf{w}}_{1}, \dots, \tilde{\mathbf{w}}_{K}]\in\mathbb{R}^{d\times K}$ and define an ETF regularizer that penalizes deviations from a simplex equiangular tight frame:
\begin{equation}
\mathcal{L}_{\mathrm{ETF}}
=
\left\|
\tilde W^{\top}\tilde W
-
\left(
\frac{K}{K-1} I
-
\frac{1}{K-1} \mathbf{1}\mathbf{1}^{\top}
\right)
\right\|_F^2.
\end{equation}
where $\mathbf{1}\mathbf{1}^{\top}$ denotes an all-ones matrix.
The overall training objective of the task-prior network is
\begin{equation}
\mathcal{L}_{\mathrm{TPN}}
=
\mathcal{L}_{\mathrm{CE}}
+
\lambda_{\mathrm{ETF}}\mathcal{L}_{\mathrm{ETF}},
\end{equation}
where $\mathcal{L}_{\mathrm{CE}}$ denotes the cross-entropy loss and $\lambda_{\mathrm{ETF}}$ controls the strength of geometric regularization. After training, the $h_{\psi}$ is frozen and applied to the unlabeled set $\mathcal{D}_{u}=\{\mathbf{X}_j\}_{j=1}^{N_u}$. For each unlabeled sample $\mathbf{X}_j$, the encoder produces an embedding$\mathbf{z}_j = f_{\psi}(\mathbf{X}_j)$, which is mapped by the linear head to task-prior logits $\mathbf{s}_j = W^{\top}\mathbf{z}_j \in \mathbb{R}^{K}.$ We then derive a prior label by $\hat{y}_j^{\mathrm{prior}} = \arg\max_{k} \; \mathbf{s}_{jk}.$

\textbf{Prototype Clustering via Intra-Class Variability.}
Given the inter-class structure induced by the TPN, we further introduce learnable prototypes to capture rich intra-class variability. For each task, we maintain a prototype bank $\mathcal{P}$ with $M$ prototypes for each class $k \in \{1,\dots,K\}$:
$
    \mathcal{P}=\{p_{k,m} \in \mathbb{R}^d | m \in \{1,2,...,M\} \}.
$
The prototype bank $\mathcal{P}$ is initialized via $k$-means clustering on unlabeled TPN embeddings grouped by predicted prior labels. For a sample $X_i$ and a prototype $p_{k,m}$, we compute their affinity using cosine similarity:
\begin{equation}
s_{i,(k,m)}
=
\left(\frac{\mathbf{z}_i}{\lVert \mathbf{z}_i \rVert_2}\right)^{\top}
\left(\frac{\mathbf{p}_{k,m}}{\lVert \mathbf{p}_{k,m} \rVert_2}\right).
\end{equation}

For class $k$, let $\mathcal{I}_k$ denote the index set of samples assigned to $k$ with $B_k = |\mathcal{I}_k|$.
We define the class-wise similarity matrix $S^{(k)} \in \mathbb{R}^{B_k \times M}$ with entries $s^{(k)}_{i,m}$.
Accordingly, a soft sample-to-prototype assignment ${Q}^{(k)}_{i,m} = exp(S^{(k)}_{i,m}/\epsilon)$ can be derived, where $\epsilon$ is a temperature parameter. we optimize $Q$ to maximize the sample-prototype similarity:
\begin{equation}
\max_{Q\in\mathcal{Q}} \ \mathrm{Tr}(Q S^\top)
= \sum_{i\in\mathcal{I}_k}\sum_{m=1}^{M} Q^{(k)}_{i,m}s^{(k)}_{i,m},
\quad
\text{s.t. } \sum_{m=1}^{M} Q^{(k)}_{i,m}=1,\ 
\sum_{i\in\mathcal{I}_k} Q^{(k)}_{i,m}=\frac{B_k}{M}.
\end{equation}
These constraints enforce a balanced assignment of samples to prototypes. we apply the Sinkhorn-Knopp algorithm to solve this optimization problem. We refine the prototype bank $\mathcal{P}$ using a cross-entropy loss on labeled samples:
\begin{equation}
\mathcal{L}_{\mathrm{P}}
=
\mathbb{E}_{k \in \mathcal{K}_{\mathrm{v}}}
\
\mathbb{E}_{i \in \mathcal{I}_k}
\
\mathrm{CE}
(
\mathbf{q}^{(k)}_{i},
\
\sigma
(
\frac{\mathbf{s}^{(k)}_i}{\epsilon}
)
),
\end{equation}
where $\mathbb{E}$ denotes empirical averaging, $\mathbf{s}^{(k)}_i, \mathbf{q}^{(k)}_{i}\in\mathbb{R}^{M}$ are the similarity vector to the class prototypes and the corresponding assignment, and $\mathcal{K}_{\mathrm{v}}=\{k \mid |\mathcal{I}_k^{\ell}|>0\}$ denotes the set of classes that appear in the batch. The prototype label for each unlabeled sample is $\hat y_j^{\mathrm{proto}}=\arg\max_{k,m} s_{j,(k,m)}$.

\subsubsection{Confidence-aware Sample-level Supervision}
\label{sec:confidence_fusion}
\textit{\textbf{Technical Challenge.}}
Direct pseudo-labeling can amplify errors under label-limited conditions, especially when confidence is estimated from a single prediction source. High-confidence logits may still reflect spurious biases rather than reliable sample-level supervision.
\\
\textit{\textbf{Design Rationale.}}
We derive pseudo-labels from the agreement between two complementary sources: task-prior predictions, which emphasize inter-class separability, and prototype-based predictions, which capture intra-class variation.

We assign a pseudo-label $\hat{y}_j$ only when the prior prediction
$\hat{y}_j^{\mathrm{prior}}$ agrees with the prototype prediction $\hat{y}_j^{\mathrm{proto}}$. We further quantify the reliability of agreed predictions using a confidence-aware fusion mechanism inspired by Dempster-Shafer theory~\citep{denoeux2000neural}.

\begin{lemma}
\label{lemma:ds_fuse}
In the singleton-only scenario and without ignorance mass, the Dempster-Shafer combination rule degenerates to $m(\{\omega_c\}) = \frac{m_1(\{\omega_c\})\, m_2(\{\omega_c\})} {\sum_{j} m_1(\{\omega_j\})\, m_2(\{\omega_j\})}$, where $m_1(\cdot)$ and $m_2(\cdot)$ denote the basic belief assignments (BBAs) provided by two evidence sources, and $\omega_c \in \Omega$ represents the $c$-th singleton hypothesis in the frame of discernment $\Omega$.
\end{lemma}

Under the setting of Lemma~\ref{lemma:ds_fuse}, we instantiate the two evidence sources as the task-prior prediction and the prototype-based prediction, respectively. Specifically, since both predictors output class-wise scores over singleton hypotheses and no explicit ignorance mass is modeled, their logits $\mathbf{s}_j$ and $\boldsymbol{\varpi}_j$ can be directly normalized to BBAs:
\begin{equation}
m_1^{(j)}(\cdot), \;m_2^{(j)}(\cdot) = \mathrm{Softmax}(\mathbf{s}_j), \; \mathrm{Softmax}(\boldsymbol{\varpi}_j),
\label{eq:bba_from_logits}
\end{equation}
The fused BBA $m^{(j)}(\cdot)$ is obtained by combining $m_1^{(j)}(\cdot)$ and $m_2^{(j)}(\cdot)$ under the singleton-only assumption without ignorance mass. The confidence of the pseudo-label is then quantified using an entropy-based confidence score:
\begin{equation}
\gamma_j
=
1
-
\frac{
-\sum_{c=1}^{K}
m^{(j)}(\{\omega_k\}) \log m^{(j)}(\{\omega_k\})
}{
\log K
}.
\label{eq:confidence}
\end{equation}
where $K$ denotes the number of classes. An unlabeled sample $\mathbf{X}_j$ is selected for adaptation only if
its confidence score $\gamma_j$ exceeds a predefined threshold $\rho$.

\subsection{Prototype-Conditioned Adaptation}
\label{sec:adapter}
\textit{\textbf{Technical Challenge.}} The key challenge is to inject prototype-level structure into EFM adaptation without overwriting pretrained representations. Prototype-based pseudo-label selection provides loss-level supervision, but cannot directly constrain intermediate feature adaptation.
\\
\textit{\textbf{Design Rationale.}} We condition lightweight adapters on sample-specific prototype similarity. This allows prototype geometry to modulate intermediate features, providing persistent structural guidance while keeping the EFM backbone frozen.

Building upon the constructed supervision,
we propose a \emph{Prototype-Conditioned Adapter (ProAdapter)}. For each unlabeled sample $\mathbf{X}_j \in \mathcal{D}_{u}$, we use its prototype similarity vector $\boldsymbol{\varpi}_j$ to condition lightweight adaptation modules inserted into a frozen EFM.

\textbf{Architecture.} \emph{ProAdapter} adopts a feature-wise modulation design to adapt intermediate representations in a parameter-efficient and architecture-agnostic manner. It operates on residual features when residual connections are available, and otherwise modulates layer outputs. In practice, \emph{ProAdapter} is inserted into the last $L$ layers of the backbone.

\textbf{Prototype-Conditioned Modulation.}
Given a prototype similarity $\boldsymbol{\varpi}_j \in \mathbb{R}^{K}$ and an intermediate representation $\mathbf{h}_l \in \mathbb{R}^{D\times T}$, \emph{ProAdapter} uses feature-wise modulation:
\begin{equation}
\mathrm{ProAdapter}(\mathbf{h}_l;\boldsymbol{\varpi}_j) =
\boldsymbol{\alpha}_j \odot \mathbf{h}_l + \boldsymbol{\beta}_j,
\label{eq:film_affine}
\end{equation}
where $\boldsymbol{\alpha}_j,\boldsymbol{\beta}_j \in \mathbb{R}^{D}$ are channel-wise scale and shift parameters, broadcast along temporal dimension. Modulation parameters are generated in two branches. First, statistics from $\mathbf{h}_l$ form a self-conditioning vector $\mathbf{c}_j = [\boldsymbol{\mu}_l, \boldsymbol{\sigma}_l]= \big[\mathrm{Mean}_t(\mathbf{h}_l),\; \mathrm{Std}_t(\mathbf{h}_l)\big] \in \mathbb{R}^{D}$ to predict base modulation parameters:
\begin{equation}
[\Delta\boldsymbol{\alpha}_j^{(0)};\boldsymbol{\beta}_j^{(0)}] = Linear(\mathbf{c}_j)\in \mathbb{R}^{2D}.
\label{eq:self_branch}
\end{equation}

In parallel, the prototype similarity $\boldsymbol{\varpi}_j$ produces an additional modulation term whose magnitude is explicitly bounded and gated to ensure stable adaptation:
\begin{equation}
\begin{aligned}
[\Delta\boldsymbol{\alpha}_j^{(\varpi)};\boldsymbol{\beta}_j^{(\varpi)}]
&= \lambda_{\varpi}\tanh\!\big(Linear(\boldsymbol{\varpi}_j)\big), 
\qquad
s_j = \sigma\!\big(Linear(\boldsymbol{\varpi}_j)\big), \\
(\Delta\boldsymbol{\alpha}_j,\boldsymbol{\beta}_j,\boldsymbol{\alpha}_j)
&= \big(
\Delta\boldsymbol{\alpha}_j^{(0)} + s_j\Delta\boldsymbol{\alpha}_j^{(\varpi)},
\boldsymbol{\beta}_j^{(0)} + s_j\boldsymbol{\beta}_j^{(\varpi)},
\mathbf{1} + \lambda\tanh(\Delta\boldsymbol{\alpha}_j)
\big).
\end{aligned}
\label{eq:proto_branch}
\end{equation}
where $\lambda_{\varpi}$ is a scaling factor. The modulated representation is then forwarded to the next layer.

\subsection{Training Strategy and Overall Objective}
\label{sec:objective}
We adopt a warm-up adaptation strategy that starts with supervised training on $\mathcal{D}_l$, and gradually incorporates unlabeled data for semi-supervised updates on $\mathcal{D}_u$. Each labeled mini-batch is paired with a fixed number of unlabeled samples to control the influence of pseudo-labels and prevent noisy supervision from dominating early training. Unlabeled data are sampled cyclically, so an epoch does not necessarily iterate over $\mathcal{D}_u$.

The semi-supervised loss is computed with sample-wise confidence weighting. For each unlabeled sample $\mathbf{X}_j \in \mathcal{D}_u$, a pseudo-label $\hat{y}_j$ and a confidence score $\gamma_j \in (0,1)$ are obtained from the supervision construction, and a combined agreement-confidence mask $m_j$ is applied. The overall training objective is
\begin{equation}
\mathcal{L}=
\mathcal{L}_{\mathrm{sup}} +
\mathbb{E}_{\mathbf{X}_j \sim \mathcal{D}_u}
\;
m_j \, \gamma_j \,
\operatorname{CE}\!\left(
\mathcal{F}_{\mathrm{EFM}}(\mathbf{X}_j),\; \hat{y}_j
\right),
\end{equation}
where $CE(\cdot,\cdot)$ denotes the cross-entropy loss, and $\mathcal{L}_{\mathrm{sup}}$ is computed on labeled samples in $\mathcal{D}_l$.

\section{Experiments}
\label{experiments}
\begin{table}[t!]
\centering
\scriptsize
\setlength{\tabcolsep}{3pt}
\begin{threeparttable}

\caption{Baseline comparisons under the limited-label setting (30\% labeled subjects).}
\label{tab:baseline_30_compact}

\begin{tabularx}{\textwidth}{l l *{12}{>{\centering\arraybackslash}X}}
\toprule
\multirow{2}{*}{Backbone} & \multirow{2}{*}{Method} & 
\multicolumn{2}{c}{\textbf{ISRUC}} & \multicolumn{2}{c}{\textbf{SEED}} &
\multicolumn{2}{c}{\textbf{Mental}} &
\multicolumn{2}{c}{\textbf{PhysioNet-MI}} &
\multicolumn{2}{c}{\textbf{SHU-MI}} &
\multicolumn{2}{c}{\textbf{CHB-MIT}} \\
\cmidrule(lr){3-4}\cmidrule(lr){5-6}\cmidrule(lr){7-8}
\cmidrule(lr){9-10}\cmidrule(lr){11-12}\cmidrule(lr){13-14}
& &
Kappa & WF1 & AUPRC & AUC & AUPRC & AUC & Kappa & WF1 & AUPRC & AUC & AUPRC & AUC \\

\midrule

\multirow{2}{*}{\textbf{Non-Pretrained}}
& EEGNet
& 45.40 & 52.02 & 53.39 & 54.15 & 40.35 & 63.75
& 34.20 & 50.82 & 51.60 & 53.02 & 16.21 & 69.26 \\

& EEGConformer
& 47.70 & 53.93 & 54.31 & 53.48 & 38.92 & 62.31
& 34.72 & 51.38 & 52.24 & 53.58 & 18.38 & 71.55 \\
\midrule

\multicolumn{14}{l}{\textit{Transformer-based backbones}} \\

\multirow{6}{*}{\textbf{LaBraM}} 
& Frozen
& 48.58 & 58.35 & 72.45 & 73.89 & 60.83 & \underline{79.25}
& 22.81 & 41.80 & 61.81 & \underline{63.49} & 30.53 & 86.30 \\

& +LoRA 
& \underline{55.61} & \underline{65.05} & 67.51 & 57.28 & \underline{64.47} & 54.16
& 22.79 & 41.69 & 61.55 & 63.12 & 32.16 & 86.96 \\

& +FixMatch \textsuperscript{\dag}
& 33.14 & 43.83 & \underline{74.08} & \textbf{77.32} & 53.92 & 74.41
& \underline{41.50} & \underline{55.64} & 49.15 & 49.62 & 29.97 & 87.00 \\

& +FineSSL \textsuperscript{\dag}
& 54.24 & 63.76 & 72.99 & 73.78 & 53.12 & 60.72
& 8.69 & 26.50 & 61.59 & 63.11 & \underline{33.66} & 86.03 \\

& +\textbf{SCOPE (Ours)} \textsuperscript{\dag}
& \cellcolor{blue!10}\textbf{56.19} & \cellcolor{blue!10}\textbf{65.15}
& \cellcolor{blue!10}\textbf{74.25} & \cellcolor{blue!10}\underline{75.23}
& \cellcolor{blue!10}\textbf{64.64} & \cellcolor{blue!10}\textbf{82.93}
& \cellcolor{blue!10}\textbf{41.80} & \cellcolor{blue!10}\textbf{56.33}
& \cellcolor{blue!10}\textbf{63.33} & \cellcolor{blue!10}\textbf{64.21}
& \cellcolor{blue!10}\textbf{36.05} & \cellcolor{blue!10}\textbf{89.13} \\

\noalign{\vskip 2pt}
\cdashline{2-14}
\noalign{\vskip 2pt}

& Full finetuning
& 51.26 & 61.73 & 66.84 & 66.25 & 45.49 & 70.72
& 38.31 & 52.73 & \underline{62.96} & 63.29 & 33.62 & \underline{87.25} \\
\midrule

\multirow{6}{*}{\textbf{CBraMod}} 
& Frozen
& 60.34 & 67.91 & 57.83 & 58.74 & 47.19 & 63.53
& 30.73 & 47.60 & 54.29 & 54.64 & 28.82 & 87.11 \\

& +LoRA 
& \underline{63.16} & \underline{70.40} & 62.59 & 62.09 & 43.14 & \underline{70.45}
& 32.97 & \underline{49.46} & 62.05 & 61.83 & 23.55 & 86.15 \\

& +FixMatch \textsuperscript{\dag}
& 20.42 & 23.23 & 60.21 & 59.16 & 46.70 & 66.85
& 23.32 & 40.15 & 51.14 & 51.04 & 30.78 & 87.53 \\

& +FineSSL \textsuperscript{\dag}
& 53.46 & 61.21 & 57.84 & 58.42 & 41.51 & 64.43
& 20.90 & 33.38 & 51.29 & 51.69 & \underline{33.64} & \underline{88.92} \\

& +\textbf{SCOPE (Ours)} \textsuperscript{\dag}
& \cellcolor{blue!10}\textbf{63.29} & \cellcolor{blue!10}\textbf{70.74}
& \cellcolor{blue!10}\textbf{65.57} & \cellcolor{blue!10}\textbf{63.71}
& \cellcolor{blue!10}\textbf{54.44} & \cellcolor{blue!10}\textbf{74.39}
& \cellcolor{blue!10}\textbf{37.84} & \cellcolor{blue!10}\textbf{53.86}
& \cellcolor{blue!10}\textbf{66.31} & \cellcolor{blue!10}\textbf{67.80}
& \cellcolor{blue!10}\textbf{35.02} & \cellcolor{blue!10}\textbf{89.04} \\

\noalign{\vskip 2pt}
\cdashline{2-14}
\noalign{\vskip 2pt}

& Full finetuning
& 61.95 & 69.78 & \underline{64.12} & \underline{63.39} & \underline{47.35} & 69.91
& \underline{33.16} & 49.31 & \underline{64.85} & \underline{66.63} & 28.59 & 85.68 \\
\midrule

\multirow{6}{*}{\textbf{CSBrain}} 
& Frozen
& 56.27 & 64.98 & 56.27 & 56.57 & 38.30 & 66.68
& 30.55 & 47.50 & \underline{58.31} & \underline{58.96} & 20.54 & 87.08 \\

& +LoRA 
& 60.30 & 68.56 & 56.98 & \underline{57.45} & 33.78 & 63.24
& \underline{37.81} & \underline{53.25} & 56.17 & 57.51 & 18.29 & 87.17 \\

& +FixMatch \textsuperscript{\dag}
& 44.53 & 53.48 & \underline{59.71} & 57.34 & \underline{43.24} & \underline{69.13}
& 27.17 & 45.49 & 51.68 & 51.48 & 32.37 & \underline{87.76} \\

& +FineSSL \textsuperscript{\dag}
& 47.65 & 54.61 & 53.46 & 53.86 & 36.41 & 66.35
& 7.05 & 22.16 & 51.61 & 52.31 & \underline{34.21} & 87.62 \\

& +\textbf{SCOPE (Ours)} \textsuperscript{\dag}
& \cellcolor{blue!10}\textbf{61.51} & \cellcolor{blue!10}\textbf{69.86}
& \cellcolor{blue!10}\textbf{63.05} & \cellcolor{blue!10}\textbf{62.09}
& \cellcolor{blue!10}\textbf{47.60} & \cellcolor{blue!10}\textbf{71.09}
& \cellcolor{blue!10}\textbf{40.62} & \cellcolor{blue!10}\textbf{55.79}
& \cellcolor{blue!10}\textbf{60.99} & \cellcolor{blue!10}\textbf{61.58}
& \cellcolor{blue!10}\textbf{34.96} & \cellcolor{blue!10}\textbf{89.21} \\

\noalign{\vskip 2pt}
\cdashline{2-14}
\noalign{\vskip 2pt}

& Full finetuning
& \underline{60.97} & \underline{69.19} & 57.54 & 57.40 & 40.08 & 67.39
& 37.33 & 53.20 & 58.02 & 58.80 & 33.90 & 86.85 \\
\midrule

\multicolumn{14}{l}{\textit{SSM-style backbones}} \\

\multirow{6}{*}{\textbf{EEGMamba}} 
& Frozen
& 60.73 & 67.94 & 58.27 & 57.30 & 37.19 & 64.98
& 30.80 & 47.90 & 56.37 & 56.92 & 30.09 & \underline{87.38} \\

& +LoRA 
& \underline{62.78} & \underline{69.79} & 58.64 & 57.35 & 38.12 & 65.45
& \underline{33.62} & \underline{50.39} & \underline{57.11} & \underline{57.35} & \underline{32.61} & 86.53 \\

& +FixMatch \textsuperscript{\dag}
& 32.00 & 44.07 & 58.86 & 57.67 & 42.61 & \underline{67.61}
& 28.76 & 45.45 & 53.38 & 53.31 & 24.82 & 85.31 \\

& +FineSSL \textsuperscript{\dag}
& 54.02 & 65.92 & 48.98 & 47.64 & 39.11 & 66.86
& 4.37 & 16.81 & 54.06 & 54.83 & 32.06 & 85.50 \\

& +\textbf{SCOPE (Ours)} \textsuperscript{\dag}
& \cellcolor{blue!10}\textbf{63.07} & \cellcolor{blue!10}\textbf{71.41}
& \cellcolor{blue!10}\textbf{60.65} & \cellcolor{blue!10}\textbf{59.02}
& \cellcolor{blue!10}\textbf{49.46} & \cellcolor{blue!10}\textbf{68.48}
& \cellcolor{blue!10}\textbf{35.75} & \cellcolor{blue!10}\textbf{52.06}
& \cellcolor{blue!10}\textbf{57.39} & \cellcolor{blue!10}\textbf{57.46}
& \cellcolor{blue!10}\textbf{33.39} & \cellcolor{blue!10}\textbf{88.94} \\

\noalign{\vskip 2pt}
\cdashline{2-14}
\noalign{\vskip 2pt}

& Full finetuning
& 62.02 & 68.71 & \underline{59.84} & \underline{57.98} & \underline{42.89} & 65.36
& 33.58 & 50.17 & 57.09 & 57.22 & 31.06 & 85.41 \\
\midrule

\multirow{6}{*}{\textbf{CodeBrain}} 
& Frozen
& 52.54 & 61.75 & 63.44 & \underline{63.93} & 46.70 & 71.25
& 26.05 & 43.80 & \underline{58.96} & 58.30 & 28.37 & 86.45 \\

& +LoRA 
& 44.08 & 53.37 & 63.83 & 58.61 & 44.05 & 68.49
& \underline{37.11} & \underline{53.09} & 58.01 & \underline{58.66} & 15.47 & 84.60 \\

& +FixMatch \textsuperscript{\dag}
& 57.23 & 66.74 & 61.58 & 60.58 & 44.53 & 69.30
& 21.01 & 40.20 & 55.52 & 54.96 & 28.81 & 85.85 \\

& +FineSSL \textsuperscript{\dag}
& 48.94 & 57.50 & 63.72 & 62.85 & \underline{47.14} & 71.25
& 10.67 & 28.87 & 57.69 & 57.60 & \underline{35.08} & \underline{87.67} \\

& +\textbf{SCOPE (Ours)} \textsuperscript{\dag}
& \cellcolor{blue!10}\textbf{63.46} & \cellcolor{blue!10}\textbf{71.11}
& \cellcolor{blue!10}\textbf{65.88} & \cellcolor{blue!10}\textbf{65.30}
& \cellcolor{blue!10}\textbf{49.30} & \cellcolor{blue!10}\textbf{72.40}
& \cellcolor{blue!10}\textbf{37.42} & \cellcolor{blue!10}\textbf{53.24}
& \cellcolor{blue!10}\textbf{60.56} & \cellcolor{blue!10}\textbf{60.32}
& \cellcolor{blue!10}\textbf{36.63} & \cellcolor{blue!10}\textbf{88.77} \\

\noalign{\vskip 2pt}
\cdashline{2-14}
\noalign{\vskip 2pt}

& Full finetuning
& \underline{58.46} & \underline{67.05} & \underline{64.30} & 63.27 & 47.10 & \underline{72.13}
& 36.05 & 51.90 & 57.46 & 58.27 & 28.33 & 87.60 \\
\bottomrule
\end{tabularx}

\begin{tablenotes}\footnotesize
\item[*] LaBraM includes SEED in pretraining. \textsuperscript{\dag} Methods also use unlabeled data for semi-supervised training.
\end{tablenotes}

\end{threeparttable}
\end{table}

\subsection{Datasets}
\label{subsection: data}
We evaluate our method on 6 public EEG datasets: \textbf{ISRUC} \citep{khalighi2016isruc} for five-class sleep staging, \textbf{SEED} \citep{zheng2015investigating} for binary emotion recognition, \textbf{Mental Arithmetic} \citep{zyma2019electroencephalograms} for binary mental workload assessment, \textbf{PhysioNet-MI} \citep{schalk2004bci2000} for four-class motor imagery, \textbf{SHU-MI} \citep{goldberger2000physiobank} for binary motor imagery, and \textbf{CHB-MIT} \citep{shoeb2009application} for binary seizure detection. For all datasets, we follow prior EFMs settings under a strict cross-subject protocol \citep{wangcbramod, zhou2025csbrain, ma2025codebrain}. We keep the original split unchanged, with 30\% of training subjects treated as labeled and the remaining subjects used as unlabeled data. Validation and test sets consist of held-out subjects unseen during training, mimicking realistic label-limited EEG adaptation scenarios.

\subsection{Experiment Settings}
\label{subsection: setting}
\textbf{Experiment Setup.} 
We evaluate our \emph{SCOPE} framework as a plug-and-play adapter on multiple pretrained EFMs. Training is performed in two stages. In the first stage, we construct supervision using only the training data: labeled samples provide direct supervision, while unlabeled samples are used without labels to induce task priors, class prototypes, and confidence-aware pseudo-labels. In the second stage, all backbone parameters of pretrained EFMs are frozen, and only the proposed \emph{ProAdapter} is optimized. Notably, the validation set is used only for model selection in the second stage and is not involved in any part of the first stage, while the test set is strictly held out throughout.

Across all backbone models, we adopt the same task-specific classifier head used in the original EFMs to ensure a fair comparison. Experiments are conducted on 40GB NVIDIA A100 GPUs. All models are trained using AdamW with a batch size of 64 and backbone-specific learning rates following its original setups. We report the mean performance over five random seeds. For all backbone models, \emph{ProAdapter} are inserted into the last three layers. We modulate transformer layer outputs and insert \emph{ProAdapter} within residual layers for SSM-style backbones.

\textbf{Backbone and Baselines.}
We evaluate the \emph{SCOPE} framework on five publicly available EFMs, spanning different architectural designs. Specifically, we consider 3 transformer-based backbones (\textbf{LaBraM} \citep{jianglarge}, \textbf{CBraMod} \citep{wangcbramod}, and \textbf{CSBrain} \citep{zhou2025csbrain}) and 2 state-space-model (SSM)-style backbones (\textbf{EEGMamba} \citep{wang2025eegmamba} and \textbf{CodeBrain} \citep{ma2025codebrain}). As baselines, we include two representative task-specific EEG models, \textbf{EEGNet} \citep{lawhern2018eegnet} and \textbf{EEGConformer} \citep{song2022eeg}. For foundation-model backbones, we report results under frozen and full fine-tuning settings as reference points, and further compare against parameter-efficient adaptation via \textbf{LoRA} \citep{hu2022lora} and two representative self-training semi-supervised frameworks, \textbf{FixMatch} \citep{sohn2020fixmatch} and \textbf{FineSSL} \citep{gan2024erasing}.

\textbf{Evaluation Metric.}
Cohen’s Kappa and weighted F1 are reported for multiclass tasks, while AUROC and AUPRC are used for binary tasks. Model selection is based on Kappa or AUROC, and the best checkpoint is used for test evaluation.

\subsection{Performance Comparison}
Table~\ref{tab:baseline_30_compact} summarizes results under the limited-label setting, where the best results are highlighted in \textbf{bold}, the second-best results are \underline{underlined}, and our method is \colorbox{blue!10}{color-coded}. Across all five backbones and six datasets, our method consistently achieves strong performance. For example, compared with frozen backbones, on ISRUC we improve Kappa by +7.61 with LaBraM and +10.92 with CodeBrain; on SEED, we improve AUPRC by +6.78 with CSBrain; and on Mental Arithmetic, we improve AUPRC by +12.27 with EEGMamba. These gains are consistently observed across Transformer-based and SSM-style backbones, indicating that \emph{SCOPE} generalizes well across different architectural families under cross-subject distribution shifts.

A closer comparison shows that full fine-tuning can underperform the frozen backbone when the backbone has already seen the target dataset during pretraining, as with LaBraM on SEED. LoRA can fall below the frozen baseline, as observed with CBraMod on ISRUC. Similarly, classical semi-supervised methods exhibit dataset- and backbone-dependent failures, such as FineSSL on PhysioNet-MI and FixMatch with the LaBraM backbone on CHB-MIT. In contrast, \emph{SCOPE} shows more stable performance across backbone-dataset combinations.

\begin{takeawaybox}
\textit{\textbf{Takeaway 1.}} Adapting EFMs under limited subjects cannot be reliably addressed by directly applying generic methods since they can be sensitive to backbone-dataset combinations.
\end{takeawaybox}

We further evaluate label-ratio sensitivity on ISRUC with all five backbones, using 5\%, 10\%, 20\%, and 50\% labeled training subjects. Detailed results are provided in Appendix. Notably, these gains are obtained with only 2-5\% trainable parameters and substantially reduced training time and GPU memory.

\begin{table}[t!]
\centering
\small
\setlength{\tabcolsep}{6pt}
\caption{Ablation on ISRUC with 30\% labels. Methods with \textsuperscript{\dag} additionally use pseudo-labeled data.}
\label{tab:ablation_isruc}
\begin{tabularx}{\textwidth}{
l
>{\centering\arraybackslash}p{1.6cm}
>{\centering\arraybackslash}p{0.8cm}
>{\centering\arraybackslash}p{1.6cm}
>{\centering\arraybackslash}p{0.8cm}
>{\centering\arraybackslash}p{1.6cm}
>{\centering\arraybackslash}p{0.8cm}
>{\centering\arraybackslash}p{1.6cm}
>{\centering\arraybackslash}p{0.8cm}
}
\toprule
\multirow{2}{*}{Method} 
& \multicolumn{4}{c}{\textbf{CSBrain} \textit{(Transformer-based)}}
& \multicolumn{4}{c}{\textbf{CodeBrain} \textit{(SSM-Style)}} \\

\cmidrule(lr){2-5}
\cmidrule(lr){6-9}

& Kappa & $\Delta$\% 
& WF1 & $\Delta$\% 
& Kappa & $\Delta$\% 
& WF1 & $\Delta$\% \\
\midrule

Frozen
& 56.27{\scriptsize$\pm$1.49} & --
& 64.98{\scriptsize$\pm$1.16} & --
& 52.54{\scriptsize$\pm$1.40} & --
& 61.75{\scriptsize$\pm$1.25} & -- \\

\rowcolor{gray!10}
& \multicolumn{8}{c}{\textit{Ablation on Supervision Construction}} \\

w/o ETF-guide \textsuperscript{\dag}
& 60.04{\scriptsize$\pm$2.39} & {\scriptsize$\downarrow$2.38\%}
& 68.15{\scriptsize$\pm$2.07} & {\scriptsize$\downarrow$2.45\%}
& 61.48{\scriptsize$\pm$1.72} & {\scriptsize$\downarrow$3.12\%}
& 68.87{\scriptsize$\pm$1.35} & {\scriptsize$\downarrow$3.15\%} \\

w/o Prototype Clustering \textsuperscript{\dag}
& 59.16{\scriptsize$\pm$1.24} & {\scriptsize$\downarrow$3.82\%}
& 67.53{\scriptsize$\pm$1.32} & {\scriptsize$\downarrow$3.33\%}
& 60.76{\scriptsize$\pm$0.94} & {\scriptsize$\downarrow$4.25\%}
& 67.91{\scriptsize$\pm$0.86} & {\scriptsize$\downarrow$4.50\%} \\

w/o Supervision construction
& 58.30{\scriptsize$\pm$1.63} & {\scriptsize$\downarrow$5.22\%}
& 66.74{\scriptsize$\pm$1.51} & {\scriptsize$\downarrow$4.47\%}
& 59.01{\scriptsize$\pm$1.45} & {\scriptsize$\downarrow$7.01\%}
& 66.78{\scriptsize$\pm$1.03} & {\scriptsize$\downarrow$6.09\%} \\

\rowcolor{gray!10}
& \multicolumn{8}{c}{\textit{Ablation on \emph{ProAdapter} Design}} \\

w/o \emph{ProAdapter} \textsuperscript{\dag}
& 58.49{\scriptsize$\pm$0.76} & {\scriptsize$\downarrow$4.91\%}
& 66.52{\scriptsize$\pm$0.78} & {\scriptsize$\downarrow$4.78\%}
& 56.96{\scriptsize$\pm$0.52} & {\scriptsize$\downarrow$10.25\%}
& 64.46{\scriptsize$\pm$0.77} & {\scriptsize$\downarrow$9.35\%} \\

w/o Confidence Weights \textsuperscript{\dag}
& 60.35{\scriptsize$\pm$0.93} & {\scriptsize$\downarrow$1.89\%}
& 67.85{\scriptsize$\pm$1.04} & {\scriptsize$\downarrow$2.88\%}
& 61.75{\scriptsize$\pm$1.18} & {\scriptsize$\downarrow$2.69\%}
& 69.22{\scriptsize$\pm$1.34} & {\scriptsize$\downarrow$2.66\%} \\

w/o Prototype Conditioning \textsuperscript{\dag}
& 58.66{\scriptsize$\pm$1.35} & {\scriptsize$\downarrow$4.63\%}
& 66.59{\scriptsize$\pm$1.10} & {\scriptsize$\downarrow$4.68\%}
& 59.60{\scriptsize$\pm$1.33} & {\scriptsize$\downarrow$6.08\%}
& 66.97{\scriptsize$\pm$1.46} & {\scriptsize$\downarrow$5.82\%} \\

\rowcolor{gray!10}
& \multicolumn{8}{c}{\textit{Ablation on Training Strategy}} \\

w/o Warm-up \textsuperscript{\dag}
& 59.29{\scriptsize$\pm$0.62} & {\scriptsize$\downarrow$3.61\%}
& 67.46{\scriptsize$\pm$0.77} & {\scriptsize$\downarrow$3.44\%}
& 61.84{\scriptsize$\pm$1.05} & {\scriptsize$\downarrow$2.55\%}
& 69.12{\scriptsize$\pm$1.26} & {\scriptsize$\downarrow$2.80\%}\\

sequential \textsuperscript{\dag}
& 59.20{\scriptsize$\pm$1.16} & {\scriptsize$\downarrow$3.76\%}
& 67.67{\scriptsize$\pm$1.21} & {\scriptsize$\downarrow$3.14\%}
& 58.28{\scriptsize$\pm$1.08} & {\scriptsize$\downarrow$8.16\%}
& 66.53{\scriptsize$\pm$1.35} & {\scriptsize$\downarrow$6.44\%}\\

Two-Stage \textsuperscript{\dag}
& 58.31{\scriptsize$\pm$0.45} & {\scriptsize$\downarrow$5.20\%}
& 66.42{\scriptsize$\pm$0.69} & {\scriptsize$\downarrow$4.93\%}
& 57.10{\scriptsize$\pm$1.29} & {\scriptsize$\downarrow$10.02\%}
& 64.60{\scriptsize$\pm$1.10} & {\scriptsize$\downarrow$9.16\%}\\

\textbf{SCOPE (Our full model)} \textsuperscript{\dag}
& \cellcolor{blue!10}\textbf{61.51}{\scriptsize$\pm$0.55} & --
& \cellcolor{blue!10}\textbf{69.86}{\scriptsize$\pm$0.69} & --
& \cellcolor{blue!10}\textbf{63.46}{\scriptsize$\pm$1.98} & --
& \cellcolor{blue!10}\textbf{71.11}{\scriptsize$\pm$0.74} & --\\
\bottomrule
\end{tabularx}
\end{table}

\subsection{Ablation Study}
We conduct ablation studies on ISRUC dataset using \emph{CSBrain} and \emph{CodeBrain} as representative backbones of different architectures. Table~\ref{tab:ablation_isruc} summarizes the results.

\textbf{1) Ablation on Supervision Construction.}
We progressively remove components from the supervision construction module. Removing ETF guidance (\textbf{w/o ETF-guide}) causes 2-3\% drops, while discarding prototype clustering (\textbf{w/o Prototype Clustering}) leads to larger degradations of 3-5\%. Disabling supervision construction (\textbf{w/o Supervision construction}) results in the largest performance loss (5-7\%), indicating that \textit{self-training is less reliable for adapting the backbone}.

\textbf{2) Ablation on \emph{ProAdapter} Design.}
Removing \emph{ProAdapter} (\textbf{w/o \emph{ProAdapter}}) reduces the method to a frozen backbone and causes the largest drops, particularly on CodeBrain (up to 9-10\%). \textit{This suggests that the adaptation effect is stronger on SSM-style backbones, possibly because residual-layer modulation provides a more effective adaptation pathway.} Removing prototype conditioning (\textbf{w/o Prototype Conditioning}) leads to 4-6\% degradation, indicating the necessity of structured supervision. Discarding confidence weighting (\textbf{w/o Confidence Weights}) results in consistent drops (2-3\%), demonstrating that suppressing unreliable samples during optimization is important.

\textbf{3) Ablation on Training Strategy.}
Removing the supervised warm-up (\textbf{w/o Warm-up}) degrades performance, showing that initialization is necessary.
The \textbf{Sequential} strategy processes labeled samples before unlabeled ones within each epoch and further reduces performance, indicating that pseudo-labeled updates require anchoring by labeled supervision.
The \textbf{Two-Stage} strategy, which trains on labeled data followed by unlabeled data only, causes the largest drop, suggesting that \textit{benefit from pseudo-labeled data is limited when supervision is fully decoupled}.

\begin{figure}[t!]
    \centering
    \includegraphics[width=1\linewidth]{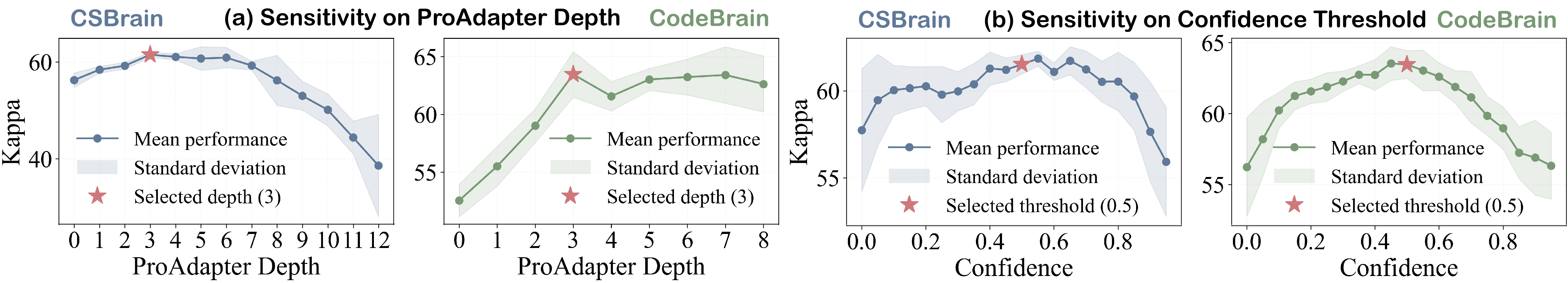}
    \caption{Sensitivity analysis on \emph{ProAdapter} depth and confidence threshold.}
    \label{fig:sen}
\end{figure}

\subsection{Sensitivity Analysis}
We analyze the sensitivity of the core design choices on ISRUC, using \emph{CSBrain} and \emph{CodeBrain} as representative backbones.

\textbf{1) Sensitivity to \emph{ProAdapter} Depth.}
Figure~\ref{fig:sen}(a) analyzes the sensitivity of \emph{ProAdapter} to the adapter depth $L$. Performance exhibits a non-monotonic trend as $L$ increases, first improving, then saturating, and potentially degrading as adaptation becomes deeper. The deeper Transformer-based \emph{CSBrain} (12 layers) exhibits sharper degradation under excessive adaptation, whereas the shallower SSM-based \emph{CodeBrain} (8 layers) remains more robust. We adopt a unified adapter depth of three layers.

\begin{takeawaybox}
\textit{\textbf{Takeaway 2:}} The optimal shallow adapter depth is consistent with our CKA analysis in \ref{fig:intro}(D), suggesting that preserving lower and middle representations while adapting task-specific top layers better balances knowledge retention and task specialization.
\end{takeawaybox}

\textbf{2) Sensitivity to Confidence Threshold.}
We analyze the sensitivity to the confidence threshold $\rho$, where only unlabeled samples with confidence greater than $\rho$ are included for adaptation. Figure~\ref{fig:sen}(b) shows that both models exhibit a clear trade-off: low thresholds admit noise and degrade performance, whereas overly high thresholds restrict supervision to only easy samples and reduce data coverage.

\section{Conclusion}
\label{Conclusion}

We present \emph{SCOPE}, a confidence-aware structured prototype-guided framework for adapting EEG foundation models under label-limited conditions. To the best of our knowledge, \textbf{this is the first systematic study on EEG foundation models adaptation under limited supervision}. Motivated by the observed failure modes of full fine-tuning, \emph{SCOPE} constructs external cohort-level structured supervision, derives confidence-aware pseudo-labels, and injects prototype information into frozen EFMs through lightweight prototype-conditioned adapters. Across diverse EEG tasks, backbone architectures, and labeled-subject ratios, \emph{SCOPE} consistently improves adaptation performance while maintaining efficiency. We hope this work provides a step toward more reliable, data-efficient, and practical deployment of EEG foundation models in real-world scenarios.

\newpage
\bibliographystyle{plainnat}
\bibliography{main}

\end{document}